\documentclass[letterpaper, 10 pt, conference]{ieeeconf}  

\usepackage{amsmath}
\usepackage{amssymb}
\usepackage{latexsym}

\usepackage{multirow}
\usepackage{booktabs} 
\usepackage{pifont}

\usepackage{hyperref}
\hypersetup{hypertex=true,
            colorlinks=true,
            linkcolor=blue,
            anchorcolor=blue,
            citecolor=blue}
\usepackage{graphicx}
\usepackage{url}
\usepackage{xcolor}
\usepackage{subfigure}
\definecolor{newcolor}{rgb}{.8,.349,.1}
\definecolor{myBlue}{rgb}{0.21,0.49,0.74}

\IEEEoverridecommandlockouts                              

\title{\LARGE \bf
Dual-Branch Graph Transformer Network for 3D Human Mesh Reconstruction from Video
}

\author{Tao Tang$^{1}$, Hong Liu$^{1*}$, Yingxuan You$^{1}$, Ti Wang$^{1}$ and Wenhao Li$^{1}$
\thanks{This paper is supported by the National Natural Science Foundation of China~(No.62373009)}
\thanks{*Corresponding authors: {\tt\small hongliu@pku.edu.cn} (Hong Liu).}
\thanks{$^{1}$Tao Tang, Hong Liu, Yingxuan You, Ti Wang and Wenhao Li are with State Key Laboratory of General Artificial Intelligence, Peking University, Shenzhen Graduate School, Shenzhen, China}%
}

\begin{document}

\maketitle
\thispagestyle{empty}
\pagestyle{empty}

\begin{abstract}
Human Mesh Reconstruction (HMR) from monocular video plays an important
role in human-robot interaction and collaboration.
However, existing video-based human mesh reconstruction methods face a trade-off between accurate reconstruction and smooth motion. These methods design networks based on either RNNs or attention mechanisms to extract local temporal correlations or global temporal dependencies, but the lack of complementary long-term information and local details limits their performance. 
To address this problem, we propose a \textbf{D}ual-branch \textbf{G}raph \textbf{T}ransformer network for 3D human mesh \textbf{R}econstruction from video, named DGTR.
DGTR employs a dual-branch network including a Global Motion Attention (GMA) branch and a Local Details Refine (LDR) branch to parallelly extract long-term dependencies and local crucial information, helping model global human motion and local human details (e.g., local motion, tiny movement).
Specifically, GMA utilizes a global transformer to model long-term human motion. LDR combines modulated graph convolutional networks and the transformer framework to aggregate local information in adjacent frames and extract crucial information of human details. Experiments demonstrate that our DGTR outperforms state-of-the-art video-based methods in reconstruction accuracy and maintains competitive motion smoothness. Moreover, DGTR utilizes fewer parameters and FLOPs, which validate the effectiveness and efficiency of the proposed DGTR. Code is publicly available at \href{https://github.com/TangTao-PKU/DGTR}{\textcolor{myBlue}{https://github.com/TangTao-PKU/DGTR}}.
\end{abstract}

\section{INTRODUCTION}
3D human mesh reconstruction is a crucial yet challenging task in computer vision and human-robot interaction \cite{iros1,iros2,iros3}, with a wide range of applications such as assisting household robots \cite{iros4} and interactive mechanical arms \cite{iros5}. 3D human mesh reconstruction is essential for higher-level intelligent human-robot interaction. These intelligent assistants should perceive the position of the human body, thereby effectively interacting with human and ensuring the safety of human-robot collaboration. 

Many methods \cite{hmr,spin,hkmr,pare,pymaf} have been proposed to recover the 3D human mesh from a single image, which doesn't require complex and expensive motion capture equipment. Simultaneously, compared with 3D human pose estimation \cite{pose1,pose2,pose3}, the human mesh can provide more information about the human body (e.g., body surface positions, body shape), which is crucial for many downstream applications (e.g., human-robot interaction, motion capture). Nevertheless, directly extracting detailed 3D human body mesh from images remains challenging due to the depth ambiguities, occlusions, and background interference. 

\begin{figure}[tbp]
\vspace{0.5em}
\centerline{\includegraphics[width=9cm]{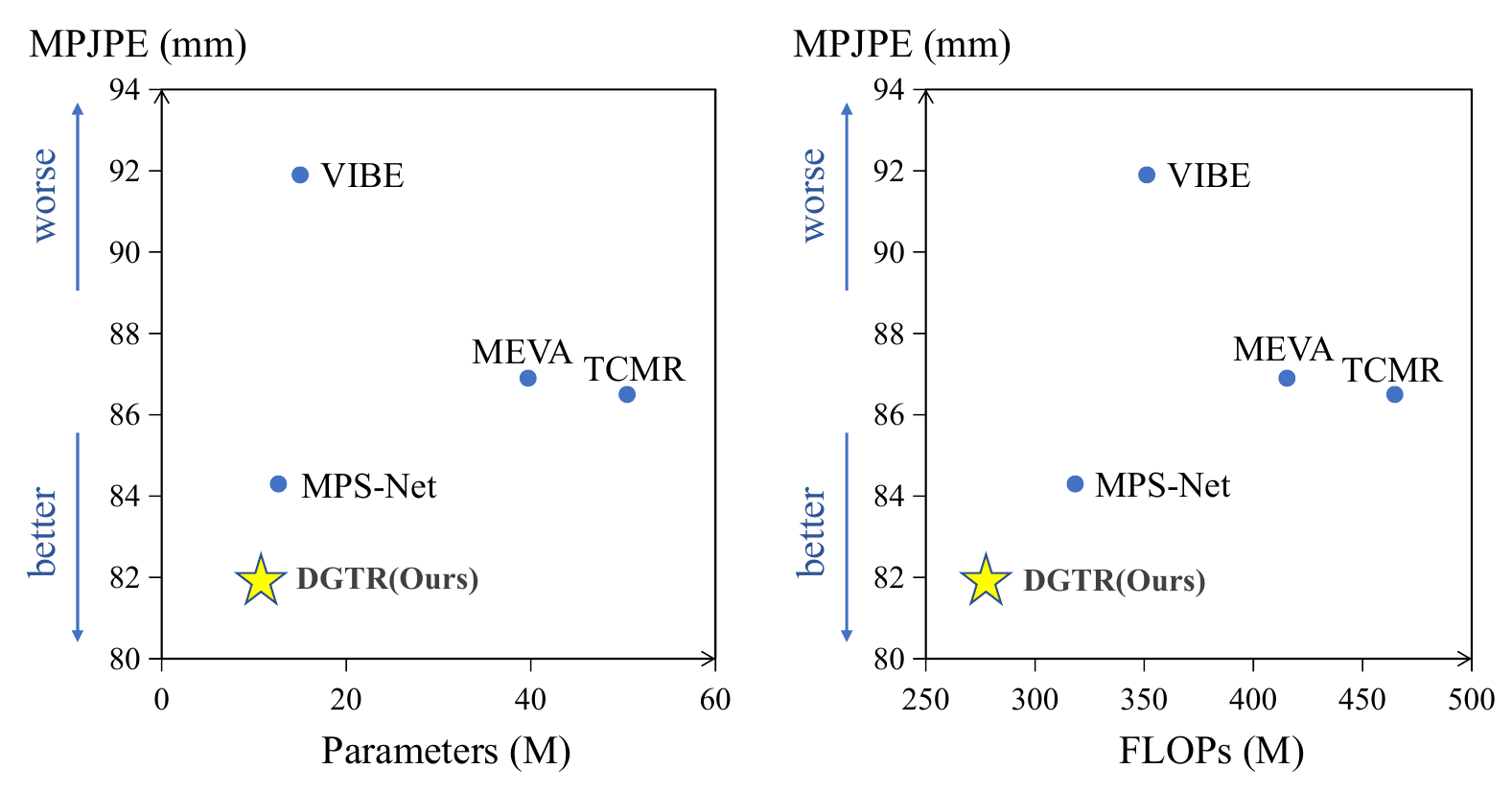}}
\caption{Comparision between accuracy (MPJPE) and parameters (left), FLOPs (right) of video-based methods. All methods are evaluated on the 3DPW dataset.}
\label{fig1}
\end{figure} 

While existing image-based methods \cite{hmr,spin,hkmr,pare,pymaf} can generate remarkably accurate 3D human mesh from individual images, they often struggle to estimate smooth 3D human pose and shape from videos due to the absence of modeling temporal continuity in human motion. To tackle this challenge, some methods \cite{hmmr,vibe,meva,tcmr,mpsnet} have extended image-based methods to video scenarios. These methods primarily leverage a pre-trained Convolutional Neural Network (CNN) to obtain the static features from images, and then serially utilize RNN-based or attention-based networks to capture temporal information. 

However, these methods have several limitations. The use of serial spatial-temporal networks in video-based methods results in a trade-off between accuracy and smoothness. RNN-based methods \cite{vibe,meva,tcmr} employs Gate Recurrent Units (GRUs) \cite{gru} to capture local temporal information, resulting in over-smoothed human motion and inaccurate human mesh. 
In contrast, the attention-based method \cite{mpsnet} effectively captures global temporal dependencies but lacks local human details.
Moreover, the coupled spatial-temporal features make it challenging for the network to balance the accuracy and smoothness of human mesh.

To address the issues above, we propose a {\bf D}ual-branch {\bf G}raph {\bf T}ransformer network for 3D human mesh {\bf R}econstruction from video~(DGTR), which parallelly handle the global temporal and local crucial information. Our method mainly consists of two branches: the Global Motion Attention (GMA) branch and the Local Details Refine~(LDR) branch. In GMA, the transformer is employed to capture global temporal information (e.g., long-term human motion). In LDR, we introduce a local information aggregation module and a crucial feature extraction module to capture local human details. 
Compared with state-of-the-art video-based method MPS-Net \cite{mpsnet}, DGTR reduces the MPJPE by 2.3mm, 2.2mm, 2.2mm on 3DPW, MPI-INF-3DHP, and Human3.6M datasets, respectively. Additionally, as shown in Fig.~\ref{fig1}, our DGTR network has fewer parameters and FLOPs among video-based methods, which is more efficient for human-robot interaction. 
Our main contributions are as follows:
\begin{itemize}
\item We propose a Dual-branch Graph Transformer network for 3D human mesh Reconstruction from video (DGTR) that parallelly captures global human motion and local human details with a dual-branch network.

\item We introduce the Global Motion Attention (GMA) branch to extract long-term human motion. Besides, we propose local information aggregation and crucial feature extraction in the Local Details Refine (LDR) branch, which aggregates local human details presented in video frames and uses Modulated GCN to capture the crucial information of local human motion.

\item We conduct extensive experiments on 3DPW, MPI-INF-3DHP and Human3.6M datasets. The results demonstrate that our DGTR surpasses previous video-based methods while using fewer parameters and FLOPs, which is efficient for practical applications.
\end{itemize}

\section{RELATED WORKS}

\subsection{Human Mesh Reconstruction}
{\bf Image-based human mesh Reconstruction.} 
Most human mesh reconstruction methods use a single image as input and regress the pose and shape parameters of the human model. 
For instance, HMR \cite{hmr} proposed an end-to-end framework that reconstructs the 3D human mesh from a single RGB image without relying on intermediate 2D key points. 
SPIN \cite{spin} proposed a self-improving network that consists of an SMPL parameter regressor and an iterative fitting framework. 
HKMR \cite{hkmr} explicitly leveraged the hierarchical structure and joint inter-dependencies of the parametric model. 
PyMAF \cite{pymaf} leveraged a feature pyramid and an explicit parameter rectification loop to improve the reliability of spatial features. 

{\bf Video-based human mesh Reconstruction.} 
In contrast to image-based methods, video-based methods need to simultaneously model accurate reconstruction and smooth human motion. 
HMMR \cite{hmmr} proposed a representation of 3D human dynamics from video sequences, enabling smooth 3D mesh motion prediction. 
VIBE \cite{vibe} employed a bidirectional GRU-based motion generator and an adversarial motion discriminator with the actual human motion dataset AMASS \cite{amass} to capture human motion. 
MEVA \cite{meva} employed a two-step encoding process involving a GRU-based motion generator and a residual refinement, effectively capturing both general human motion and person-specific details. 
TCMR \cite{tcmr} proposed a three-branch temporal encoder that effectively utilizes temporal information from past and future frames with GRUs to constrain the target frame. 
MPS-Net \cite{mpsnet} replaced GRU with an attention module to capture non-local temporal features, which can learn global and long-term human motion. 
However, these RNN-based methods \cite{vibe,meva,tcmr} still suffer from inadequate temporal information extracted by GRUs, resulting in motion jitter and inaccurate mesh estimation. Although attention-based MPS-Net effectively extracts long-term temporal information, the lack of local human details results in insufficient human details. 
Therefore, we find that the lack of complementary global motion information or local human details limits their performance, which leads to the trade-off between the accuracy and smoothness of human mesh reconstruction. 

\begin{figure*}[hbtp]
\includegraphics[width=\textwidth]{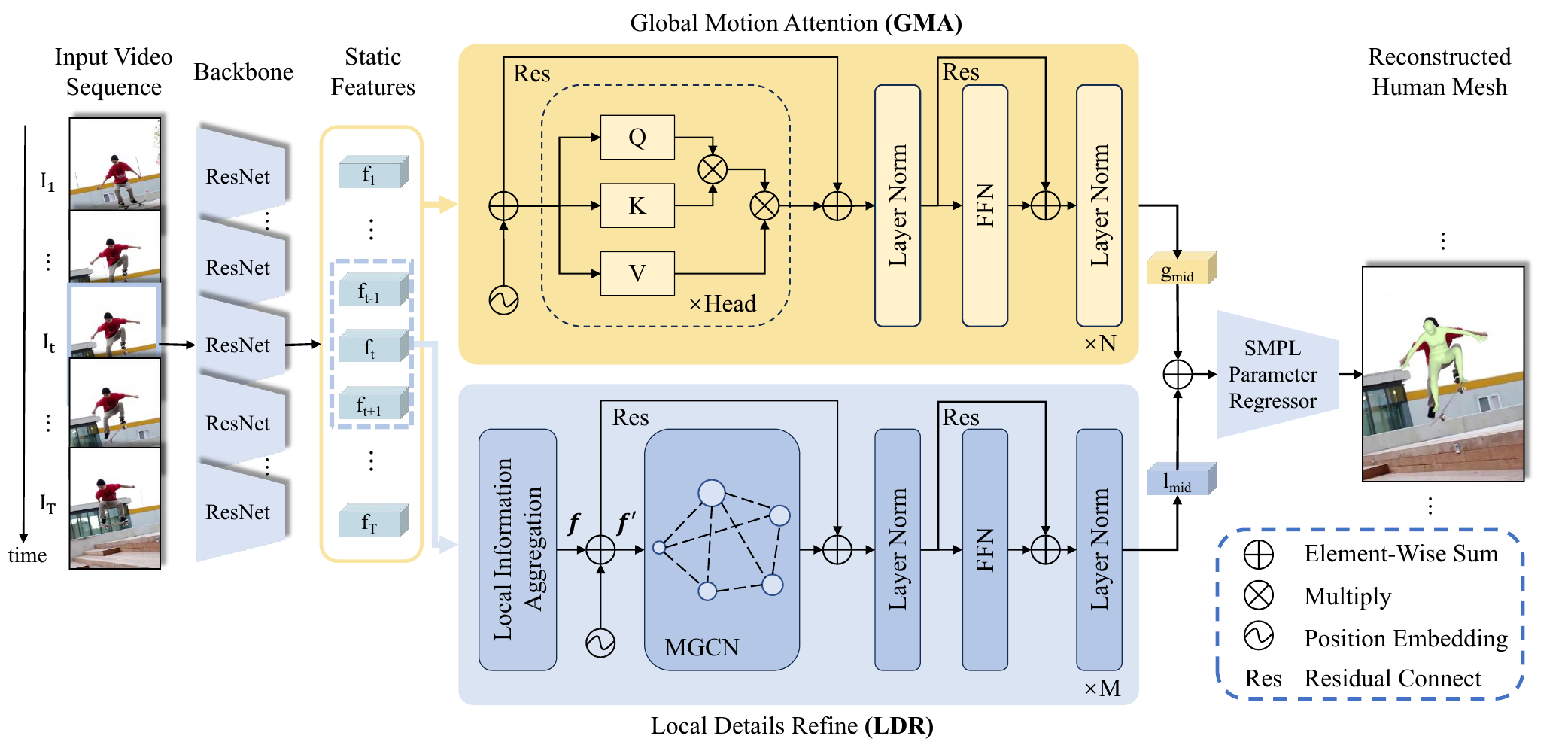}
\caption{The overview of dual-branch graph transformer network for 3D human mesh reconstruction from video (DGTR). Given a video sequence, ResNet \cite{resnet} is utilized to extract the static features. The static features of all frames and adjacent $3$ frames are separately fed into the GMA and LDR network. Then, GMA extracts global human motion, and local details of human motion are obtained by LDR. Finally, DGTR adds the output of the GMA and LDR branch and feeds it to the SMPL parameter regressor to generate the specific human mesh.} \label{fig2}
\end{figure*}

\subsection{Graph Convolutional Networks}
Graph Convolutional Networks (GCNs) \cite{gcn} generalize the capabilities of CNNs by performing convolution operations on graph-structured data. 
Recently, GCNs have been widely applied to 3D human pose and shape reconstruction \cite{graphcmr,pose2mesh,gator,igaNet}, primarily for extracting and integrating information from various human joints or meshes. 
GraphCMR \cite{graphcmr} utilized a Graph-CNN to encode the template mesh structure and process image-based features on the mesh, which enables it to regress the 3D locations of mesh vertices directly rather than predicting the model parameters (SMPL \cite{smpl}). 
Pose2Mesh \cite{pose2mesh} employed GCNs to directly estimate the 3D coordinates of human mesh vertices from the 2D human pose, addressing appearance domain gap issues and challenging pose estimation. 
Modulated GCN \cite{mgcn} disentangled feature transformations for different body joints through weight modulation and extends the graph structure beyond the human skeleton using affinity modulation while keeping relatively small parameters.

\subsection{Vision Transformer}
Transformer \cite{transformer} is first proposed for Natural Language Processing (NLP) tasks and rapidly outperforms CNN-based and RNN-based methods in many other tasks.  
Vision Transformer (ViT) \cite{vit} first replaced convolutional architecture with the transformer in the image classification task. Moreover, many researchers \cite{metro,gtrs} have employed transformers in the human mesh reconstruction task. 
For instance, METRO \cite{metro} utilized an image-based transformer encoder to jointly model vertex-vertex and vertex-joint interactions for the first time. GTRS \cite{gtrs} employed a pose-based graph transformer to exploit structured and implicit joint correlations. 
Metaformer \cite{metaformer} introduced a general architecture abstracted from the transformer without specifying the token mixer and could achieve competitive performance. 
Motivated by previous works, we leverage a transformer to extract global temporal features among video frames and employ GCNs based on the transformer framework to capture local crucial features of human motion in the video-based 3D human mesh reconstruction task.

\section{METHOD}
\subsection{Overview of DGTR}\label{AA}
The framework of DGTR is shown in Fig.~\ref{fig2}. Given a video sequence 
$V=\left\{{{I_{t}}}\right\}_{t=1}^{T}$ with $T$ frames, we firstly use ResNet \cite{resnet} pretrained by SPIN \cite{spin} to extract static feature sequence 
$F=\left\{{{f_{t}}}\right\} _{t=1}^{T}$ of all frames, where $f_{t}\in \mathbb{R}^{2048}$. 
We regard $(\left\lfloor T/2\right \rfloor+1)_{th}$ frame as the target frame, $1,...,(\left\lfloor T/2\right \rfloor)_{th}$ frames as past frames and $(\left\lfloor T/2\right \rfloor + 2)_{th},...,T_{th}$ frames as future frames. 
Then, we feed the static feature of all frames and adjacent $3$ frames separately into the Global Motion Attention (GMA) branch and the Local Details Refine (LDR) branch in parallel. 
These branches extract global temporal features and local fine-grained details, respectively. We add the global motion feature $g_{mid}$ and local refined feature $l_{mid}$ at the mid-frame, where $g_{mid}\in \mathbb{R}^{2048}$, $l_{mid}\in \mathbb{R}^{2048}$. Finally, the combined information is fed into the pre-trained SMPL Parameter Regressor \cite{spin} to obtain the vertices of the human mesh. We introduce each branch in DGTR as follows.

\subsection{Global Motion Attention}\label{AA}
RNN-based networks like GRU often struggle to capture long-term dependencies adequately and may become trapped in local temporal modeling. On the contrary, the transformer architecture has gained a significant reputation for effectively capturing long-term temporal dependencies in various tasks owing to its multi-head attention mechanism and robust global information retention capabilities. Therefore, the transformer is an ideal choice for capturing global human motion across video frames.

As shown in Fig.~\ref{fig2}, we employ the transformer encoder to extract global information from all adjacent 16 frames, facilitating the generation of smooth human motion. Specifically, we employ only $N=2$ layers of the transformer encoder block encoding temporal feature, which makes the network remain lightweight. The $\mathbb{R}^{2048}$ static features of each frame are used as input tokens and the learnable position encoding is added. We set the heads of multi-head attention mechanisms to $Head=8$ and the hidden dimension $d$ to $512$, which can be formulated as follows:
\begin{equation}
\text{Attention}(Q,K,V)=\text{Softmax}(QK^{T}/\sqrt{d})V.
\label{eq0}
\end{equation}
Finally, we utilize the output features $g_{mid}\in \mathbb{R}^{2048}$ corresponding to the target frame as the output of this branch.
\subsection{Local Details Refine}\label{AA}
{\bf Local Information Aggregation.} 
The input of LDR is a static feature sequence 
$F=\left\{{{f_{t}}}\right\} _{t=1}^{T}\in \mathbb{R}^{T\times2048}$ of nearby $3$ frames.
Due to the use of pooling operations in ResNet50 \cite{resnet} for feature extraction from images, which loses a significant portion of the information relevant to the human body in the images. This harms the representation capability of static features and makes it extremely challenging to recover complex human body mesh from the static features.
\begin{table*}[tbph]
\caption{Evaluation of State-of-the-art methods on 3DPW, MPI-INF-3DHP, and Human3.6M datasets. All methods use 3DPW for training, but do not use the SMPL parameters of Human3.6M from Mosh \cite{mosh}. The top two best results are highlighted in \textbf{bold} and \underline{underlined}, respectively.}
\begin{center}
\small
\renewcommand\arraystretch{1.25}  
\setlength{\tabcolsep}{0.3mm}{
\begin{tabular}{l|cccc|ccc|ccc|c}
\toprule
\multirow{2}{*}{Method} & \multicolumn{4}{c|}{3DPW}                                                                         & \multicolumn{3}{c|}{MPI-INF-3DHP}                                          & \multicolumn{3}{c|}{Human3.6M}                                       & \multirow{2}{*}{\begin{tabular}[c]{@{}c@{}}input\\ frames\end{tabular}} \\  \cline{2-11} 
                        & {\scriptsize PA-MPJPE~$\downarrow$} 
                        & {\scriptsize MPJPE~$\downarrow$} 
                        & {\scriptsize MPVPE~$\downarrow$} 
                        & {\scriptsize ACC-ERR~$\downarrow$} 
                        & {\scriptsize PA-MPJPE~$\downarrow$} 
                        & {\scriptsize MPJPE~$\downarrow$} 
                        & {\scriptsize ACC-ERR~$\downarrow$} 
                        & {\scriptsize PA-MPJPE~$\downarrow$} 
                        & {\scriptsize MPJPE~$\downarrow$} 
                        & {\scriptsize ACC-ERR~$\downarrow$} &                                                                         \\ \midrule
HMMR (CVPR'19) \cite{hmmr}                    & {72.6}     & {116.5} & {139.3} & 15.2    & {-}        & {-}     & -       & {56.9}     & {-}     & -       & \textbf{16}                                                                     \\ 
VIBE (CVPR'20) \cite{vibe}                    & {57.6}     & {91.9}  & {-}     & 25.4    & {68.9}     & {103.9} & 27.3    & {53.3}     & {78.0}  & 27.3    & \textbf{16}                                                                      \\ 
MEVA (ACCV'20) \cite{meva}                    & {54.7}     & {86.9}  & {-}     & 11.6    & {65.4}     & \underline{96.4}  & 11.1    & {53.2}     & {76.0}  & 15.3    & 90                                                                      \\ 
TCMR (CVPR'21) \cite{tcmr}                    & {52.7}     & {86.5}  & {102.9} & \textbf{6.8}     & {63.5}     & {{97.3}}  & \textbf{8.5}     & {52.0}     & {73.6}  & {3.9}     & \textbf{16}                                                                      \\ 
MPS-Net (CVPR'22) \cite{mpsnet}        & {52.1}     & {84.3}  & {99.7} & {7.4}     & {62.8}     & {96.7}  & {9.6}    & {\underline{47.4}}     & {\underline{69.4}}  & \textbf{3.6}  & \textbf{16}   \\ 
Zhang et. al (CVPR'23) \cite{zhang}        & {\underline{51.7}}     & {\underline{83.4}}  & {\underline{98.9}} & \underline{7.2}     & {\underline{62.5}}     & {98.2}  & \underline{8.6}    & {51.0}     & {73.2}  & \textbf{3.6}  & \textbf{16}    
\\ 
DGTR~(Ours)         & {\textbf{51.3}}     & {\textbf{82.0}}  & {\textbf{97.3}} & {7.6}     & {\textbf{61.3}}     & \textbf{94.5}  & \textbf{8.5}    & {\textbf{46.1}}     & {\textbf{67.2}}  & \underline{3.8}  & \textbf{16}    
\\ \bottomrule
\end{tabular}}
\label{table:tab1}
\end{center}
\end{table*}
To address this problem, we introduce the local information aggregation module. This module leverages 1D convolution to aggregate the local human details among adjacent $3$ frames in the static feature sequence. 
It effectively utilizes local redundancy in video frames to model local human details (e.g., local motion, tiny movement). 
This is achieved by sliding a small kernel along the input sequence and the weights within the kernel are learned during training. This process enhances the capacity of the network to focus on the crucial human details in images.

{\bf Crucial Feature Extraction.} 
Following local information aggregation, we obtain a local crucial feature sequence. Subsequently, we leverage a crucial feature extraction module to extract crucial human details (e.g., local motion, tiny movement) from the nearby frames related to the target frame.
Inspired by MetaFormer \cite{metaformer} and the good crucial feature extraction capability of the transformer framework, we introduce a novel module by integrating Modulated GCN \cite{mgcn} into the transformer framework. 
Modulated GCN \cite{mgcn} has demonstrated its effectiveness in capturing local crucial features of the human body with two components: weight modulation and adjacency modulation. Weight modulation utilizes modulation vectors to adjust or modulate the shared weight matrix for each node individually while maintaining small model parameters. On the other hand, adjacency modulation overcomes the limitation of relying on a pre-defined adjacency matrix to propagate and fuse information among all nodes. Modulated GCN appends a learnable adjacency matrix to capture strong correlations between long-term video frames. Modulated GCN can be expressed as:
\begin{equation}
Y=\text{sigmoid}(D^{-\frac{1}{2} } \tilde{A} D^{-\frac{1}{2}} X (W\odot V)),
\label{eq1}
\end{equation}
where $X$ represents the input of all nodes in graph, $\tilde{A}$ donates a learnable adjacency matrix, $D$ is a degree matrix, $W$ is a fusion matrix that needs to be learned, $V$ represents modulation vectors, and $Y$ is the output of Moudulated GCN.

As shown in Fig.~\ref{fig2}, we replace the multi-head attention mechanism with Modulated GCN in the transformer encoder, which enables us to effectively enforce local constraints while significantly reducing network parameters and FLOPs. Specifically, each video frame is treated as a graph node, and the adjacency matrix of the graph is initialized as all ones and can be learned through extensive training on mixed 3D and 2D human datasets. The specifics of the crucial feature extraction module can be calculated as follows:
\begin{equation}
f^{'} =f+PE,
\label{eq4}
\end{equation}
\begin{equation}
m=\text{Norm}(\text{MGCN}(f^{'})),\label{eq5}
\end{equation}
\begin{equation}
l_{mid}=\text{Norm}(\text{FFN}(m)+m),\label{eq6}
\end{equation}
where $PE$ represents the  position embedding of each frame, $\text{Norm}$ denotes the layer normalization, $\text{MGCN}$ denotes the Modulated GCN, $\text{FFN}$ denotes feedforward networks. Besides, We only employ $M=1$ layer of the LDA module.

\subsection{Loss Function}\label{AA}
Following previous methods \cite{vibe,tcmr,mpsnet}, we apply SMPL shape loss, SMPL pose loss, 3D joint location loss, and 2D joint reprojection loss to help the network learn the accurate human body from images. 
To further constrain the network to generate smooth human motion, we employ two additional losses to utilize the velocity of the predicted 3D/2D joint locations. More details about the loss function are available on our project page.

\section{EXPERIMENTS}
\subsection{Implementation Details}
Consistent with the previous methods \cite{vibe,tcmr,mpsnet}, we set the input sequence length $T$ to 16. We utilize the pre-trained ResNet50 from SPIN \cite{spin} to extract static features relevant to the human body of each frame. The SMPL parameter regressor consists of a single fully connected layer with 1024 neurons, followed by an output layer that predicts SMPL pose, shape, and camera parameters. 
We initialize the learning rate to $1\times10^{-4}$ and use the batch size of 64. Additionally, we employ a Cosine Annealing scheduler with the linear warm-up for the Adam optimizer \cite{adam}. We train the entire network for 50 epochs, all experiments are conducted on a single NVIDIA GTX 3090 GPU. 
PyTorch is utilized for code implementation.
\begin{table}[tbp]
\caption{Comparison of the parameters and FLOPs.}
\begin{center}
\small
\renewcommand\arraystretch{1.25}  
\setlength{\tabcolsep}{3.0mm}{
\begin{tabular}{l|c|c}
\toprule
Method & Parameters~(M) & FLOPs~(M)\\ 
\midrule
VIBE \cite{vibe}& 15.01	&  351.19 \\
MEVA \cite{meva}  & 39.70&  415.43 \\
TCMR \cite{tcmr} &50.43&  464.80 \\
MPS-Net \cite{mpsnet} &12.65&  318.39 \\
DGTR~(Ours)& \textbf{10.89}&  \textbf{277.56} \\
\bottomrule
\end{tabular}}
\label{table:tab2}
\end{center}
\end{table}

\begin{table}[tbp]
\caption{Ablation results for different branches of DGTR on 3DPW dataset.}
\begin{center}
\small
\renewcommand\arraystretch{1.25}  
\setlength{\tabcolsep}{0.3mm}{
\begin{tabular}{l|cccc}
\toprule
\multirow{2}{*}{Method}
& \multicolumn{4}{c}{3DPW}                                                                         \\ \cline{2-5} 
                  & \multicolumn{1}{c}{PA-MPJPE~$\downarrow$} & \multicolumn{1}{c}{MPJPE~$\downarrow$} & \multicolumn{1}{c}{MPVPE~$\downarrow$} & ACC-ERR~$\downarrow$ \\ \midrule
 \ {DGTR wo. GMA}                    & \multicolumn{1}{c}{52.3}     & \multicolumn{1}{c}{84.4}  & \multicolumn{1}{c}{99.7} & 7.9    \\ 
\ {DGTR wo. LDR}                    & \multicolumn{1}{c}{52.0}     & \multicolumn{1}{c}{82.6}  & \multicolumn{1}{c}{98.8} & \textbf{7.6}    \\ 
\ {DGTR}                    & \textbf{51.3}     & \textbf{82.0}  & \textbf{97.3} & \textbf{7.6}    \\ 
 \bottomrule
\end{tabular}
}
\label{table:tab3}
\end{center}
\end{table}
\subsection{Datasets and Evaluation metrics}
{\bf Datasets.} 
Following previous methods \cite{vibe,tcmr,mpsnet}, we train our model on 3DPW \cite{3dpw}, Human3.6M \cite{h36m}, MPI-INF-3DHP \cite{mpii3d}, and InstaVariety \cite{hmmr}. Subsequently, we evaluate the performance of our model on the 3DPW, Human3.6M, and MPI-INF-3DHP datasets that include 3D joint annotations. 

{\bf Evaluation metrics.} We utilize metrics for both accuracy and smoothness. To evaluate reconstruction accuracy, we employ the mean per joint position error (MPJPE), Procrustes-aligned MPJPE (PA-MPJPE), and mean per vertex position error (MPVPE). These metrics measure the difference between predicted position and ground truth in millimeters $(mm)$. 
To evaluate reconstruction smoothness, we utilize the acceleration error (ACC-ERR) proposed in HMMR \cite{hmmr}. This metric calculates the average difference in acceleration of body joints, which is measured in $(mm/s^{2})$.

\subsection{Comparison with state-of-the-art methods}
{\bf Comparison with video-based methods.} 
As shown in Table \ref{table:tab1}, we conduct extensive experiments to compare our DGTR with previous video-based methods \cite{hmmr,vibe,meva,tcmr,mpsnet,zhang} that report acceleration error. 
The results show that our DGTR outperforms state-of-the-art methods in almost all metrics, achieving the best reconstruction accuracy and competitive motion smoothness. 
This demonstrates that our dual-branch graph transformer network is proficient in concurrently modeling global dependencies (e.g., long-term human motion) and capturing local details (e.g., local motion, tiny movement).
\begin{figure}[tbhp]
\vspace{0.5em}
\centerline{\includegraphics[width=9cm]{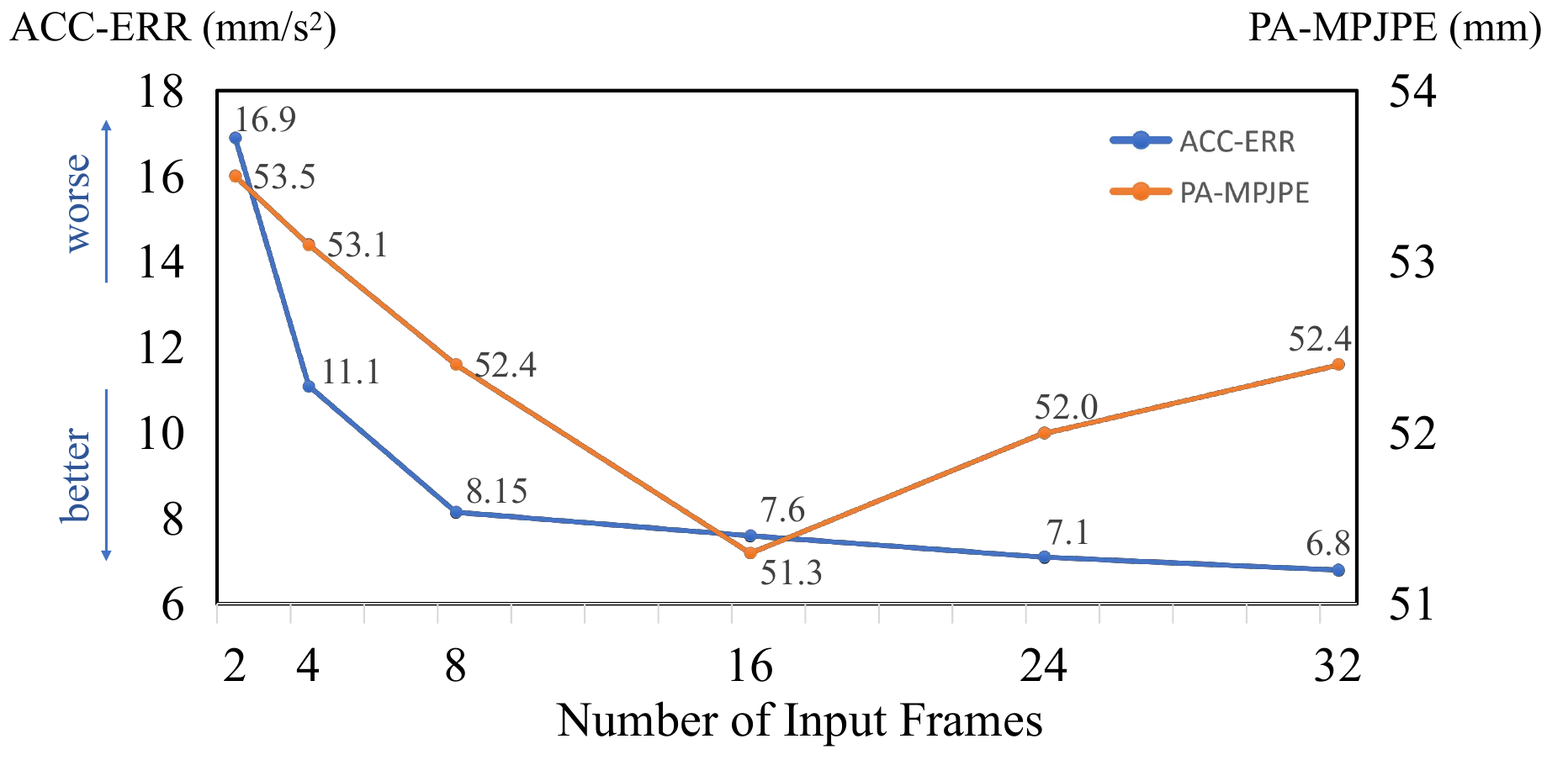}}
\caption{Different number of input frames of DGTR under acceleration error and PA-MPJPE.}
\label{fig5}
\end{figure}
Since Zhang et al.\cite{zhang} didn’t release their code, making it impossible to compare qualitative experimental results and the parameters of their model. Therefore, we primarily compare our mothed with MPS-Net\cite{mpsnet}.
In terms of reconstruction accuracy, DGTR achieves the most accurate human mesh reconstruction compared with the previous methods. Compared to MPS-Net \cite{mpsnet}, our DGTR improves MPJPE by 2.3$mm$ (from 84.3$mm$ to 82.0$mm$), 2.2$mm$ (from 96.7$mm$ to 94.5$mm$) and 2.2$mm$ (from 69.4$mm$ to 67.2$mm$) on the 3DPW, MPI-INF-3DHP, and Human3.6M datasets, respectively.  
The dual-branch network significantly improves the accuracy of reconstruction. The local details refine branch efficiently aggregates local details among adjacent frames, emphasizing the local information of human details. 
Regarding the smoothness of human motion, our DGTR achieves the best acceleration error on the MPI-INF-3DHP dataset and competitive results on the other two datasets. 
Although attention-based MPS-Net achieves similar motion smoothness to our DGTR, the reliability of global attention modules and the absence of local human details result in inaccurate reconstruction. 
On the contrary, our dual-branch DGTR alleviates the trade-off between smoothness and accuracy, achieving the best performance among video-based methods.

{\bf Comparison in parameters and FLOPs.} 
As shown in Table \ref{table:tab2}, taking the example of reconstructing one frame with an input of $T=16$ frames. 
Compared to previous video-based methods, our DGTR achieves fewer parameters and FLOPs while outperforming previous methods in almost all metrics.

\begin{figure*}[tbp]
\vspace{-4em}
\includegraphics[width=\textwidth]{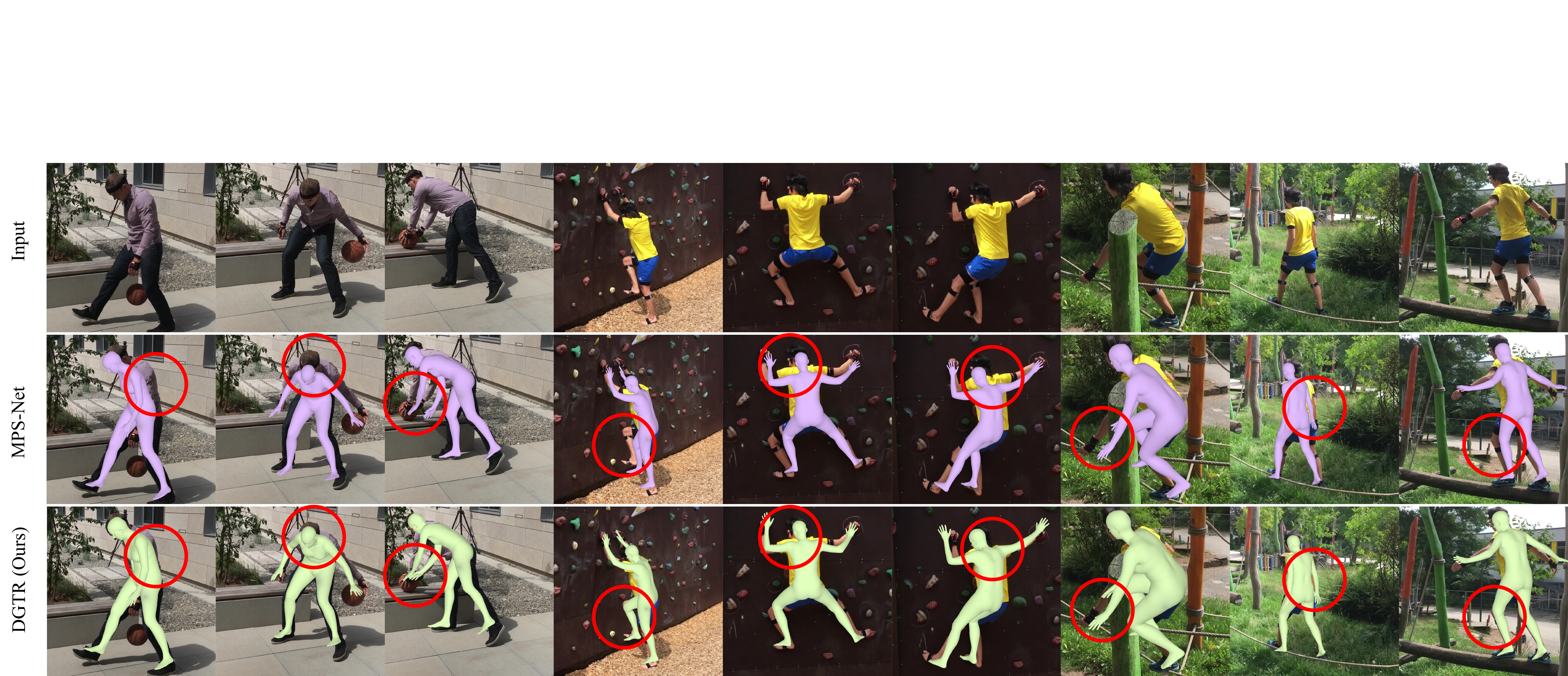}
\caption{Qualitative comparison of MPS-Net and DGTR on 3DPW dataset. Including fast motion, self-occlusion, and object occlusion scenarios.} \label{fig6}
\end{figure*}

\begin{figure}[tbhp]
\centerline{\includegraphics[width=9cm]{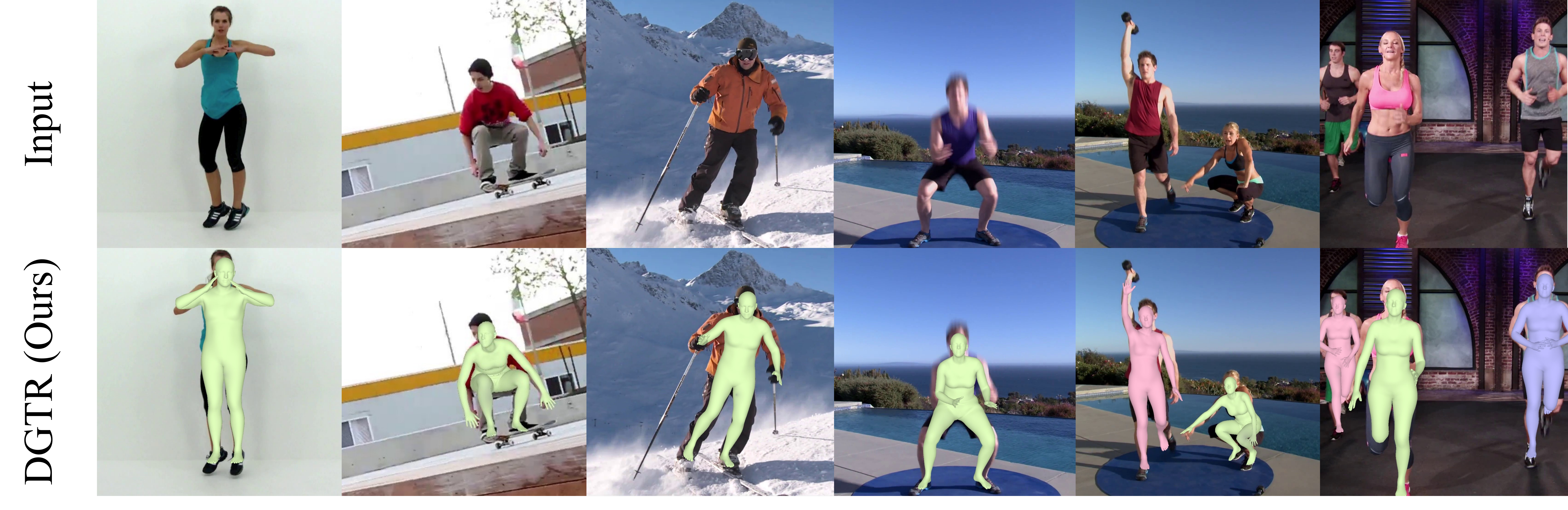}}
\caption{Qualitative results of DGTR on Internet video. Including complex backgrounds, motion blur, and multi-person scenarios.}
\label{fig7}
\end{figure} 

\subsection{Ablation Analysis}
{\bf Effectiveness of GMA and LDR branch.} 
As shown in Table \ref{table:tab3}, 
we conduct experiments to validate the effectiveness of each branch of DGTR. We evaluate the model on the outdoor 3DPW dataset. 
Experimental results demonstrate that the performance of only using the LDR branch is limited due to the utilization of only one layer of GCN, resulting in insufficient capacity to fit the high-dimensional mapping from images to the human mesh. 
When only the GMA branch is employed, the model achieves great motion smoothness. However, similar to MPS-Net \cite{mpsnet}, 
the lack of local human details leads to inaccurate reconstruction.
The DGTR achieves the best performance by incorporating the LDR branch and the GMA branch.

{\bf Impact of sequence lengths.} 
To investigate the impact of different numbers of input frames on the performance of DGTR, we conduct ablation experiments by setting the input length to 2, 4, 8, 16, 24, and 32. As shown in Fig.~\ref{fig5}, 
the results demonstrate that the accuracy tends to decrease as the input sequence length surpasses 16 frames. 
The transformer captures excessive temporal information from distant frames, leading to a reduction in single-frame reconstruction accuracy. 
Regarding the smoothness of human motion, the smoothness is improved as the input sequence length increases. 
This can be attributed to the fact that the transformer benefits from longer input sequences by acquiring more temporal information, which enhances temporal consistency in the motion. 
To ensure a fair comparison with previous video-based methods \cite{hmmr,vibe,tcmr,mpsnet}, we set the input sequence length to 16.

\subsection{Qualitative Evaluation}
{\bf Comparison on experimental dataset.} 
As shown in Fig.~\ref{fig6}, compared to MPS-Net \cite{mpsnet}, DGTR achieves more accurate human body reconstruction in fast basketball playing, slow climbing motion, and outdoor sports with multiple occlusions. As shown in the $7_{th}$ column of Fig.~\ref{fig6}, the parallel dual-branch DGTR can more effectively utilize global and local information under slight occlusion, which reconstructs better human details.

{\bf Visualization on Internet videos.} 
To evaluate the generalization capability of DGTR, we employ our model on videos from the Internet with various motions and scenes. 
As shown in Fig.~\ref{fig7}, the results demonstrate that our DGTR can generate accurate human mesh and smooth human motion in single-human,  motion blur, and multi-human scenes. 
\begin{figure}[tbhp]
\vspace{-0.5em}
\centerline{\includegraphics[width=9cm]{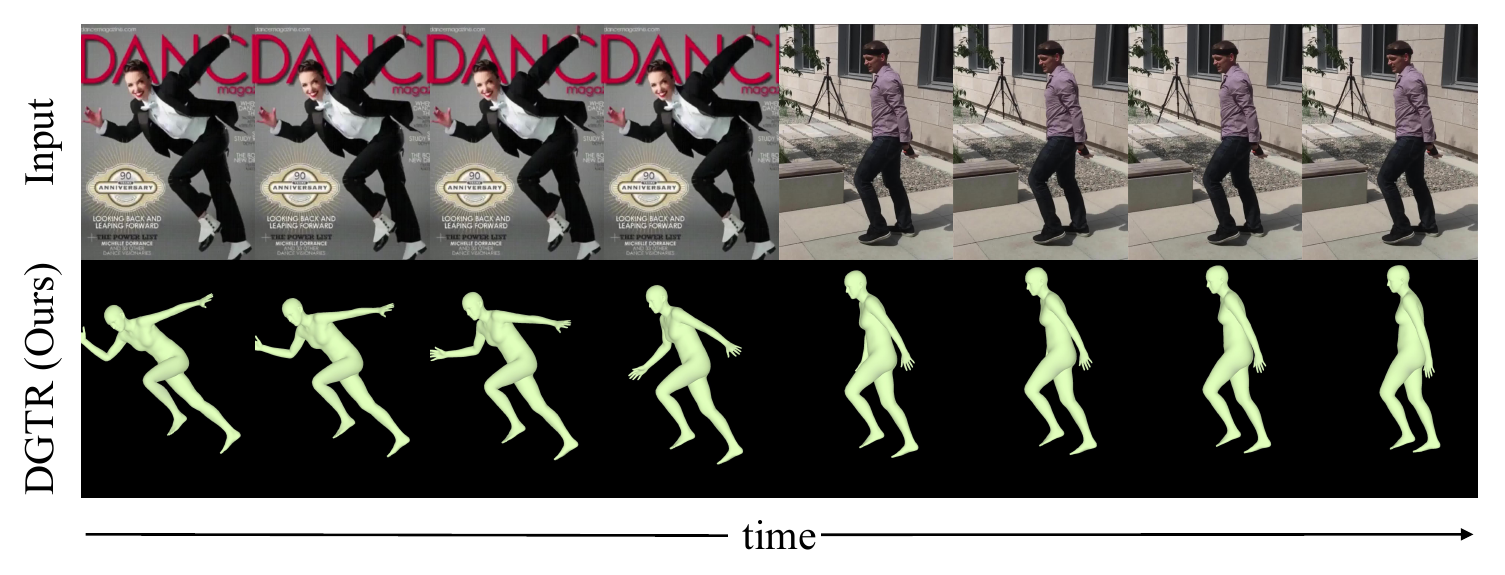}}
\vspace{-0.5em}
\caption{Qualitative results of DGTR on a stitched video.}
\vspace{0.4em}
\label{fig8}
\end{figure} 

{\bf Analysis on a stitched video.} Qualitative evaluation above mainly focuses on per-frame reconstructed accuracy. To demonstrate the robust temporal feature extraction and motion jitter reduction capability of our DGTR, we conduct experiments on a stitched video. We use two entirely different images and repeated them 30 times to create a stitched video. 
As shown in Fig.~\ref{fig8}, frames $27_{th}$ to $34_{th}$ are displayed. 
Although the input human motion suddenly changes between $30_{th}$ and $31_{st}$ frames, we can note that the motion of the arms and legs changes gradually over time in the reconstructed human mesh.

\section{CONCLUSIONS}
In this paper, we present DGTR, a Dual-branch Graph Transformer network for 3D human mesh Reconstruction from videos.
We introduce a novel network with two branches: Global Motion Attention (GMA) and Local Details Refine (LDR), which can parallelly model global human motion and local human details. The GMA branch utilizes the transformer encoder to model long-term temporal information (e.g., long-term human motion). The LDR branch introduces a novel CNN-based information aggregation module and GCN-based transformer framework, which can effectively capture crucial information of human details (e.g., local motion, tiny movement). Compared with state-of-the-art methods, our DGTR achieves the best reconstruction accuracy and competitive motion smoothness while using fewer parameters and FLOPs. 
Moreover, experiments verify the efficiency of our DGTR for human mesh reconstruction, which shows its potential for practical applications.

{\bf Limitation.} Main limitation of our method comes from severe occlusion and truncation of human body.







\newpage
\bibliographystyle{./IEEEtran}
\bibliography{final}

\end{document}